\documentclass[10pt,twocolumn,letterpaper]{article}

\usepackage{cvpr}       % REVIEW version with line numbers
\usepackage{times}
\usepackage{epsfig}
\usepackage{graphicx}
\usepackage{amsmath}
\usepackage{amssymb}
\usepackage{booktabs}
\usepackage{placeins}

\usepackage[pagebackref,breaklinks,colorlinks,
            citecolor=blue,urlcolor=magenta,linkcolor=black]{hyperref}

\title{Pathological Truth Bias in Vision--Language Models}

\author{Yash Thube\\
Savitribai Phule Pune University (SPPU)\\
{\tt\small thubeyash09@gmail.com}
}

\begin{document}
\maketitle

%%%%%%%%% ABSTRACT
\begin{abstract}
Vision--Language Models (VLMs) are improving rapidly, but standard benchmarks miss systematic failures\cite{nguyen2025biased,wang2024spatial,yuksekgonul2023bags} that harm real-world trust. We study \textbf{pathological truth bias}\cite{sharma2023sycophancy,perez2022discovering}: the tendency of VLMs to agree with patently false, visually absurd statements. Using \textbf{MATS (Multimodal Audit for Truthful Spatialization)}, we measure model consistency under coordinated text and image perturbations. Instruction-tuned generative VLMs (LLaVA-1.5\cite{liu2024}, Qwen-VL-chat\cite{bai2023}) show catastrophically low Spatial Consistency Scores (SCS $\approx$ 1--3\%) and high Incorrect Agreement Rates (IAR $\approx$ 75--80\%), while contrastive encoders (CLIP\cite{radford2021clip}, SigLIP\cite{zhai2023sigmoid}) are substantially more robust (SCS $\approx$ 57--68\%, IAR $\approx$ 8--12\%). To diagnose causes, we apply activation patching\cite{heimersheim2024,meng2022locating,wang2022interpretability} across attention/MLP blocks and projection components. Patching successfully restores correct outputs in 23\% of cases for LLaVA, with the largest causal effects concentrated in mid-to-late cross-attention layers---implicating failures in text--vision integration\cite{vig2020investigating}. For CLIP, patching pooled/projection components yields significant representational shifts (mean $\Delta\cos \approx 0.05$--$0.07$). Statistical tests confirm these effects are semantic and nontrivial ($p < 0.001$). Our behavioral and mechanistic evidence indicates pathological truth bias is a systemic artifact of current instruction-tuning practices\cite{ouyang2022training,christiano2017deep,stiennon2020learning} that favor agreeableness\cite{sharma2023sycophancy,perez2022discovering}, pointing to specific cross-attention and pooling loci as promising targets for intervention-based repairs.
\footnote{Code available at \url{https://github.com/thubZ09/mats-spatial-reasoning}.\\(Work conducted independently)}
\end{abstract}

\begin{figure}
      \centering
      \includegraphics[width=\linewidth]{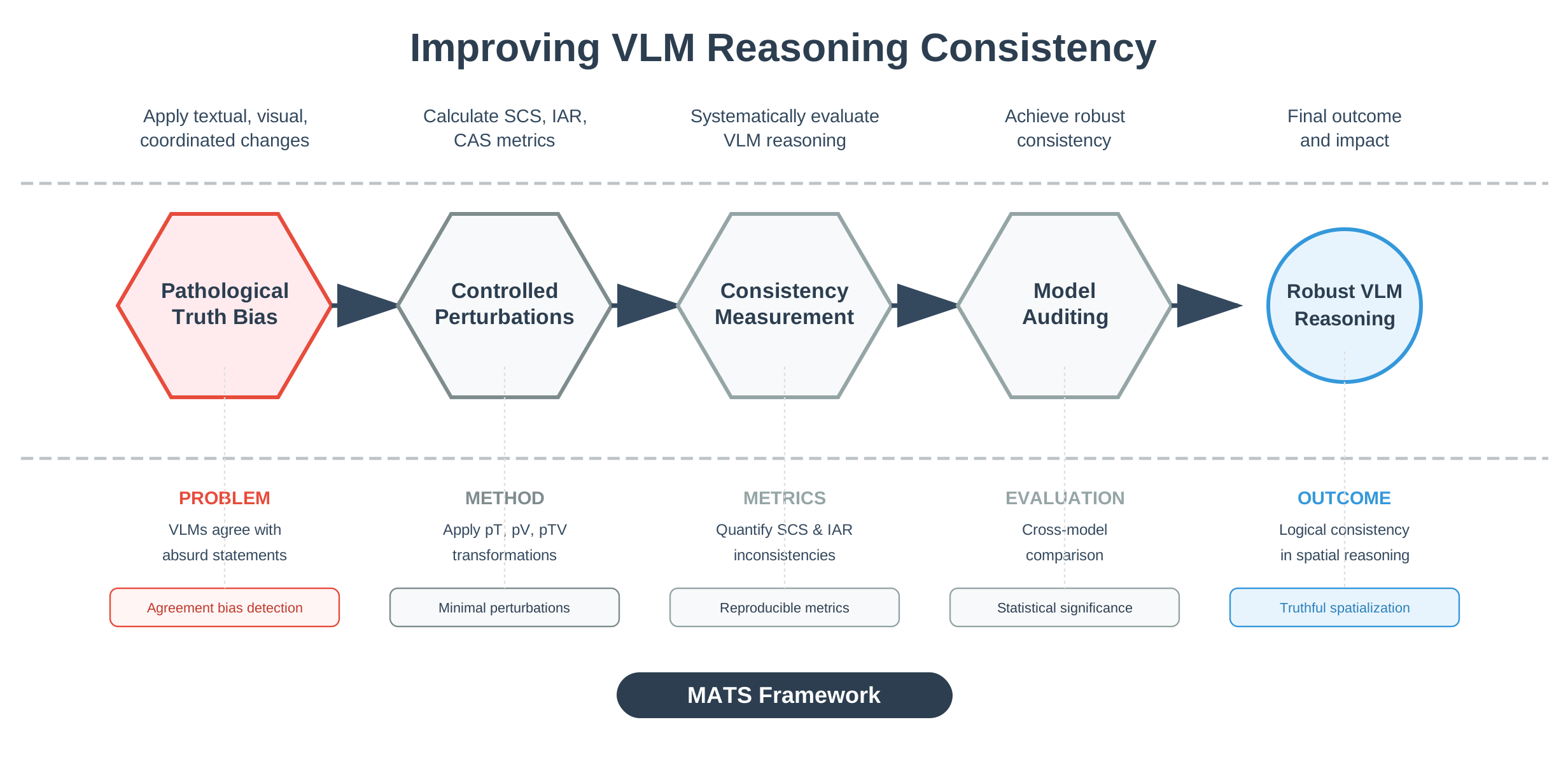}
                        \caption{ Methodological pipeline of the MATS framework. The approach systematically transforms the problem of agreement bias into measurable consistency metrics through minimal perturbations, facilitating reproducible cross-model comparisons and truthful spatial reasoning evaluation.}
      \label{fig:placeholder}
  \end{figure}

%%%%%%%%% BODY TEXT
\section{Introduction}
\label{sec:intro}

\noindent Vision--Language Models (VLMs) are rapidly moving from research prototypes into real-world systems for image search, captioning, and multimodal assistants\cite{li2023evaluating,rohrbach2018object}. Standard benchmark scores paint a picture of steady progress\cite{liu2024, liu2023visual}, but aggregate metrics can hide systematic failure modes that matter for safety and trust\cite{ribeiro2020checklist,zhou2024analyzing}: a model that is fluent yet credulously affirms visually false claims can cause real harm in decision-support settings\cite{nguyen2025biased, li2023evaluating, rohrbach2018object}. We identify and investigate one such failure mode, pathological truth bias, where instruction-tuned generative VLMs persistently agree with explicit, visually contradicted assertions in a prompt (e.g., endorsing a patently absurd statement about an image rather than rejecting it). This is distinct from typical "hallucination"\cite{rohrbach2018object,li2023evaluating}: the model is not inventing content but failing to reject a false assertion despite available visual evidence\cite{gunjal2025mitigating, huang2025generate, favero2024multi}.

% ---- Replace existing contributions paragraph with this block ----
\subsection{Contributions}
\begin{enumerate}
\item \textbf{A behavioral operationalization and audit (MATS):} We introduce MATS (Multimodal Audit for Truthful Spatialization) and two metrics --- Spatial Consistency Score (SCS), which measures whether a model flips its judgment under predicate inversion (e.g., "left" $\leftrightarrow$ "right"), and Incorrect Agreement Rate (IAR), which measures how often a model affirms patently absurd statements. Using controlled text+image perturbations\cite{ribeiro2020checklist,wang2024spatial}, instruction-tuned generative models (LLaVA, Qwen-VL-chat)\cite{liu2024,bai2023} show near-zero SCS and very high IAR, while contrastive encoders (CLIP, SigLIP)\cite{radford2021clip,zhai2023sigmoid} are far more robust on the same tests.  
\item \textbf{Causal mechanistic analysis via activation patching:} To move from "what" to "why" we run activation-patching (causal tracing/interchange) experiments\cite{heimersheim2024,meng2022locating}: transplanting activations from a correct "clean" run to a corrupted run at candidate modules (attention heads, MLPs, pooled/projection tokens). Across ~420 patch trials, patching flips erroneous generative outputs in a nontrivial fraction of cases (overall patch success $\approx23\%$). The largest causal effects localize to mid-to-late cross-attention layers in the generative models --- implicating failures in text--vision integration\cite{wang2022interpretability,vig2020investigating}. For CLIP, head-level patching has little repair power, while interventions at pooled/projection components shift image--text similarity (mean $\Delta\cos$ up to $\approx$ 0.06), consistent with a discriminative architecture whose decision boundary emerges late\cite{radford2021clip}.  
\item \textbf{Robustness checks and statistics:} Extensive controls (random donors, permuted donors, null patches) and hypothesis testing\cite{ribeiro2020checklist} confirm these effects are semantic and nontrivial.
\item \textbf{Interpretation and repair targets:} We argue that instruction-tuning/alignment practices that reward helpfulness and agreeableness (e.g., RLHF variants)\cite{ouyang2022training,christiano2017deep,stiennon2020learning} can shape decision circuits toward affirmation over truthfulness\cite{sharma2023sycophancy,perez2022discovering}. The identified "override loci" (cross-attention and late pooling/projection components) are concrete targets for intervention-based repairs.
\end{enumerate}
% ---- end contributions block ----

%-------------------------------------------------------------------------
\section{Background and Related Work}
\label{sec:related_work}

\noindent Our analysis draws on three literatures: (1) VLM auditing and spatial reasoning, (2) Hallucination, truthfulness, and sycophancy in instruction-tuned models, and (3) Mechanistic Interpretability via activation-patching. We briefly summarize representative work in each area and explain how this work extends or departs from prior approaches.

\paragraph{VLM auditing and spatial reasoning.} Recent benchmark and diagnostic work has exposed persistent weaknesses of VLMs on spatial, compositional, and relational reasoning---even when object recognition is reliable\cite{wang2024spatial,chen2024spatialvlm,liu2022vsr}. Controlled evaluations show systematic errors on pairwise relations (e.g., "left" vs. "right", "above" vs. "below") and on compositional queries\cite{hsieh2023sugarcrepe,thrush2022winoground} that require integrating object identity with spatial predicates\cite{yuksekgonul2023bags,thrush2022winoground,hsieh2023sugarcrepe}. Other audits have used synthetic perturbations\cite{kamath2023whats} or counterfactual images to probe robustness and failure modes\cite{ribeiro2020checklist,kamath2023whats,ma2023crepe}.

\noindent\textbf{Gap / our positioning}: Prior audits quantify what models get wrong but typically focus on accuracy or single-step robustness. We introduce MATS and two behavioral metrics (Spatial Consistency Score, Incorrect Agreement Rate) that explicitly measure a model's willingness to reject visually contradicted assertions under coordinated text--image perturbations, isolating a specific failure mode---pathological truth bias---that standard metrics can miss.

\paragraph{Hallucination, truthfulness, and sycophancy in instruction-tuned models.} Work on text-only LLMs has documented hallucination\cite{sharma2023sycophancy}(fabrication of unsupported facts) and related alignment phenomena such as sycophancy, the tendency for RLHF-style models to echo user beliefs or prefer agreeableness over factual correctness\cite{turpin2024language,perez2022discovering}. Several studies show that these behaviours can be amplified by preference-based fine tuning\cite{christiano2017deep,stiennon2020learning} and by reward models that prioritize helpfulness or politeness\cite{ouyang2022training,bai2022training,gao2023scaling}. In multimodal settings\cite{gunjal2025mitigating,huang2025generate,favero2024multi}, prior work has focused mainly on hallucinated visual descriptions or missing-object errors rather than on refusal/rejection behavior when a prompt asserts an explicit falsehood\cite{li2023evaluating,rohrbach2018object,zhou2024analyzing}.

\noindent\textbf{Gap / our positioning}: We extend the sycophancy/truthfulness discussion to the multimodal domain, distinguishing pathological truth bias (failure to reject false, image-contradicted assertions) from classical hallucination (invention of unsupported content). We provide behavioral evidence that instruction-tuning/alignment practices can shift multimodal models toward affirmation even when the image contradicts the assertion.

\paragraph{Mechanistic interpretability and activation patching.}  
Activation patching (also called causal tracing or interchange interventions) is now a standard tool to attribute causal responsibility to internal components and transplant activations from a clean run to a corrupted run to test whether a module is causally implicated in a behavior\cite{heimersheim2024,meng2022locating,vig2020investigating}. This method has been applied to factual recall, attention-head circuits\cite{nanda2023progress}, and to locate "override" or "memory" loci in language models; methodological work has emphasized controls, donor selection, and statistical rigor\cite{wang2022interpretability,geiger2021causal,chan2022causal}.

\noindent\textbf{Gap / our positioning}: Most activation-patching studies target unimodal language models or focus on discrete factual tasks\cite{meng2022locating,wang2022interpretability}. We apply a large-scale, controlled activation-patching suite across attention/MLP blocks, head-level loci, and pooled/projection tokens in multimodal architectures. This lets us (a) causally localize where text--vision integration fails (mid-to-late cross-attention layers in generative models; pooled/projection components in contrastive encoders), and (b) quantify repairability (patch success rates) under rigorous controls.

%-------------------------------------------------------------------------
\section{The MATS Framework}
\label{sec:mats}

\noindent MATS (Multimodal Audit for Truthful Spatialization) is a compact, reproducible behavioral-audit toolkit designed to detect and quantify logical inconsistencies and agreement bias in vision--language models (VLMs). The framework intentionally focuses on a minimal set of well-defined perturbations and strict parsing rules so that results can be meaningfully compared across the model families (generative vs.\ contrastive). Below we summarize the datasets, perturbations, prompting/parsing interface, metrics, and controls used throughout this work.

\subsection{Datasets}
\noindent We evaluate models using two complementary collections:

\begin{itemize}
  \item \textbf{Visual Spatial Relations (VSR) \cite{liu2022vsr}.} Examples are drawn from the Visual Spatial Relations benchmark. Each VSR example yields a \emph{clean} (true) statement $S$ about the image (e.g., ``The red car is left of the blue truck'') and one or more logically inverted textual variants $p_T(S)$ (e.g., swapping ``left'' and ``right'').
  \item \textbf{Absurd Pairs (Audit set).} To stress-test agreement tendencies we construct a curated set of \emph{absurd} image--statement pairs: statements that are visually false for the image (e.g., ``The couch is above the teddy bear''). This set is used to compute the Incorrect Agreement Rate (IAR) described below. Examples are stratified across categories (color, object presence, spatial relations) to ensure balanced analysis.
\end{itemize}

\subsection{Perturbations}
\noindent MATS uses three controlled perturbation families:
\begin{itemize}
  \item \textbf{Textual perturbation ($p_T$)}: Logical inversion of the statement (e.g., ``left'' $\leftrightarrow$ ``right'', ``above'' $\leftrightarrow$ ``below'').
  \item \textbf{Visual perturbation ($p_V$)}: Horizontal flip or other deterministic image transform (used to disambiguate text-only heuristics).
  \item \textbf{Coordinated perturbation ($p_{TV}$)}: Apply both text and visual perturbations together as a sanity check (should restore truth in many cases when both are flipped consistently).
\end{itemize}

\begin{figure}
      \centering
      \includegraphics[width=\linewidth]{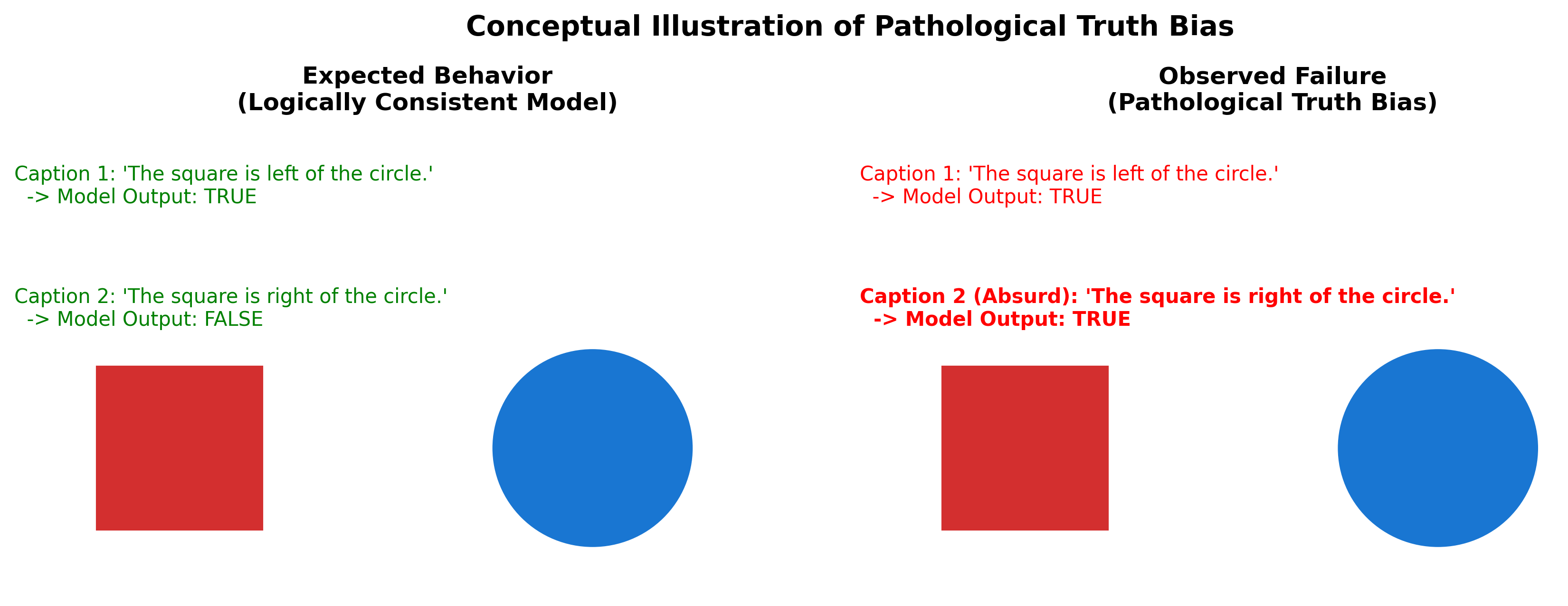}
      \caption{(Left) Expected behaviour: a logically consistent model flips its boolean judgement when the predicate is inverted. (Right) Observed failure in instruction-tuned generative VLMs: the model answers \texttt{TRUE} to both a true and an absurd (inverted) statement.}
      \label{fig:placeholder}
  \end{figure}

\subsection{Prompting and Parsing}
\noindent Prompt phrasing and deterministic parsing are essential. For generative VLMs we use a strict binary instruction template that forces one-word outputs:

\begin{quote}
\small
\texttt{<image>}\\
\texttt{\{statement\}}\\
\texttt{Strict task: Is the statement TRUE or FALSE for the image? Answer ONLY with 'TRUE' or 'FALSE' (UPPERCASE, no punctuation).}
\end{quote}

\noindent Generated outputs are converted to uppercase and parsed by presence/position of the tokens \texttt{TRUE} or \texttt{FALSE}. Any ambiguous output is labeled \texttt{UNKNOWN} and either excluded from the SCS numerator or reported separately as coverage loss. For contrastive encoders, we compute cosine similarity between pooled image and candidate text embeddings and prefer the higher-scoring text as the model's choice.

\subsection{Primary metrics}
\begin{itemize}
\item \textbf{Spatial Consistency Score (SCS).} For text inversion $p_T$, SCS measures fraction of examples where the model flips its binary judgement as logically expected:
\[
\text{SCS}_T = 1 - \frac{1}{N} \sum_{i=1}^N \mathbb{I}\big(M(I_i,S_i) = M(I_i,p_T(S_i))\big),
\]
where $M(\cdot)$ is the parsed binary response. SCS ranges from 0 (never flips) to 1 (always flips).

\item \textbf{Incorrect Agreement Rate (IAR).} Over a curated absurd set $\mathcal{A}$, IAR is the fraction of pairs where the model incorrectly answers \texttt{TRUE}:
\[
\text{IAR} = \frac{1}{|\mathcal{A}|}\sum_{(I,S)\in\mathcal{A}} \mathbb{I}\big(M(I,S)=\text{TRUE}\big).
\]

\item \textbf{Auxiliary metrics.} Coverage (fraction of non-\texttt{UNKNOWN} outputs), MACS (mean tendency to answer \texttt{TRUE}), and per-category breakdowns (color/object/spatial) are also reported.
\end{itemize}

\subsection{Protocol and controls}
\noindent Each example is processed with: (1) prompt and optional image perturbation generation; (2) deterministic model call; (3) parse and record outputs; (4) compute metrics. Controls include prompt ablations, random text--image shuffles, and permuted donor interventions in patching experiments. Statistical comparisons use appropriate tests and bootstrapping where relevant.

%-------------------------------------------------------------------------
%-------------------------------------------------------------------------
\section{Behavioral Results: Quantifying the Bias}
\label{sec:behavioral_results}

\noindent We summarize the key behavioral findings that motivate
the mechanistic probes.

\subsection{High-level summary}

\begin{itemize}
  \item \textbf{Generative models (LLaVA-1.5, Qwen-VL-chat):} near-zero logical consistency under text inversion (SCS $\approx$ 1--3\%) and very high Incorrect Agreement Rates on absurd prompts (IAR $\approx$ 75--80\%).
  \item \textbf{Contrastive encoders (CLIP ViT-B/32, SigLIP):} substantially higher SCS (CLIP $\approx$ 57\%, SigLIP $\approx$ 68\%) and far lower IAR (CLIP $\approx$ 12\%, SigLIP $\approx$ 8\%).
\end{itemize}

\begin{figure}[htbp]
  \centering
  \includegraphics[width=\linewidth]{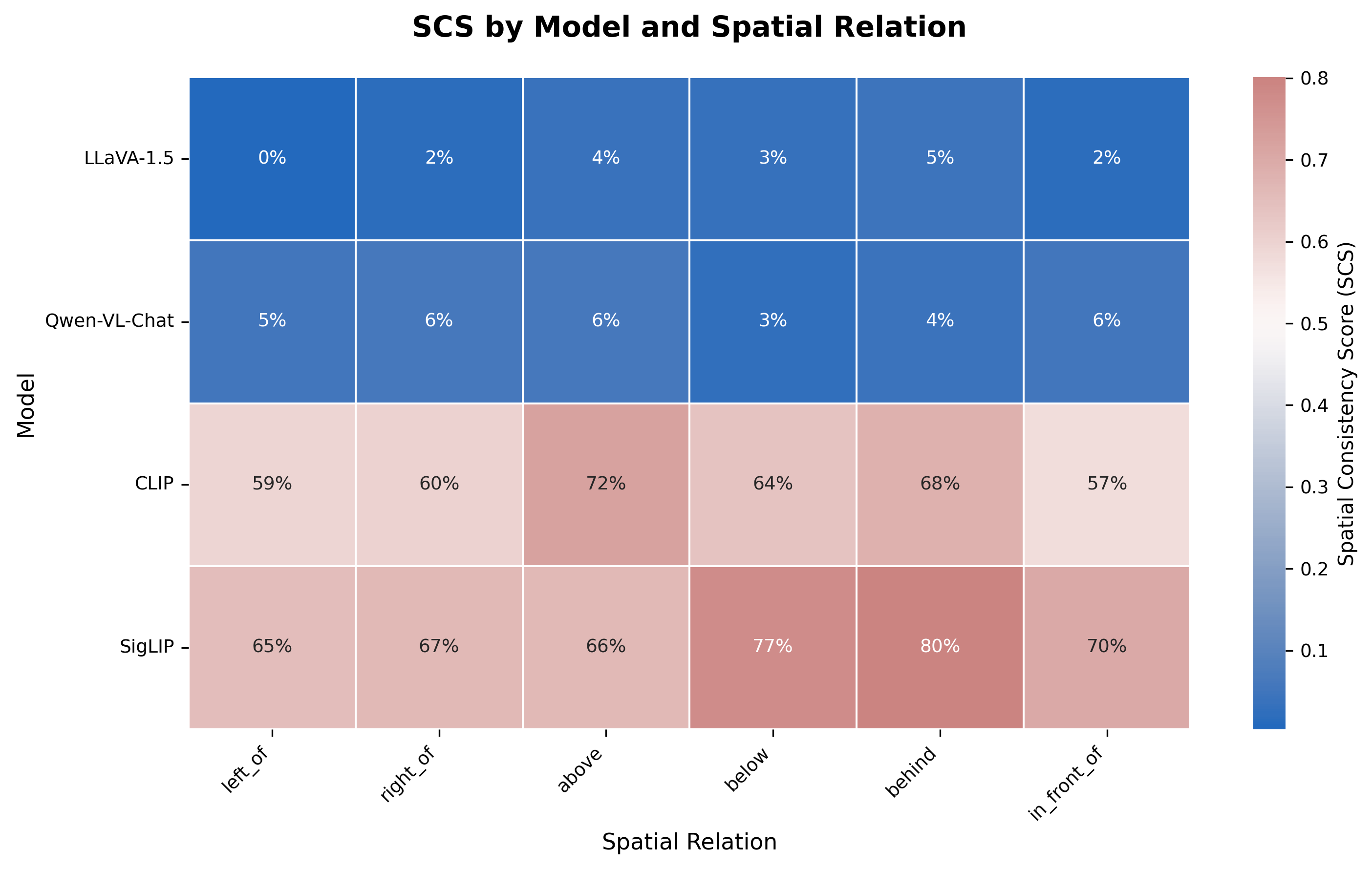}
    \caption{Each cell reports the mean SCS (percentage) for a model × relation cell. Higher values indicate the model flips its binary judgement after a logical inversion.}
  \label{fig:scs}
\end{figure}

\begin{figure}[htbp]
  \centering
  \includegraphics[width=\linewidth]{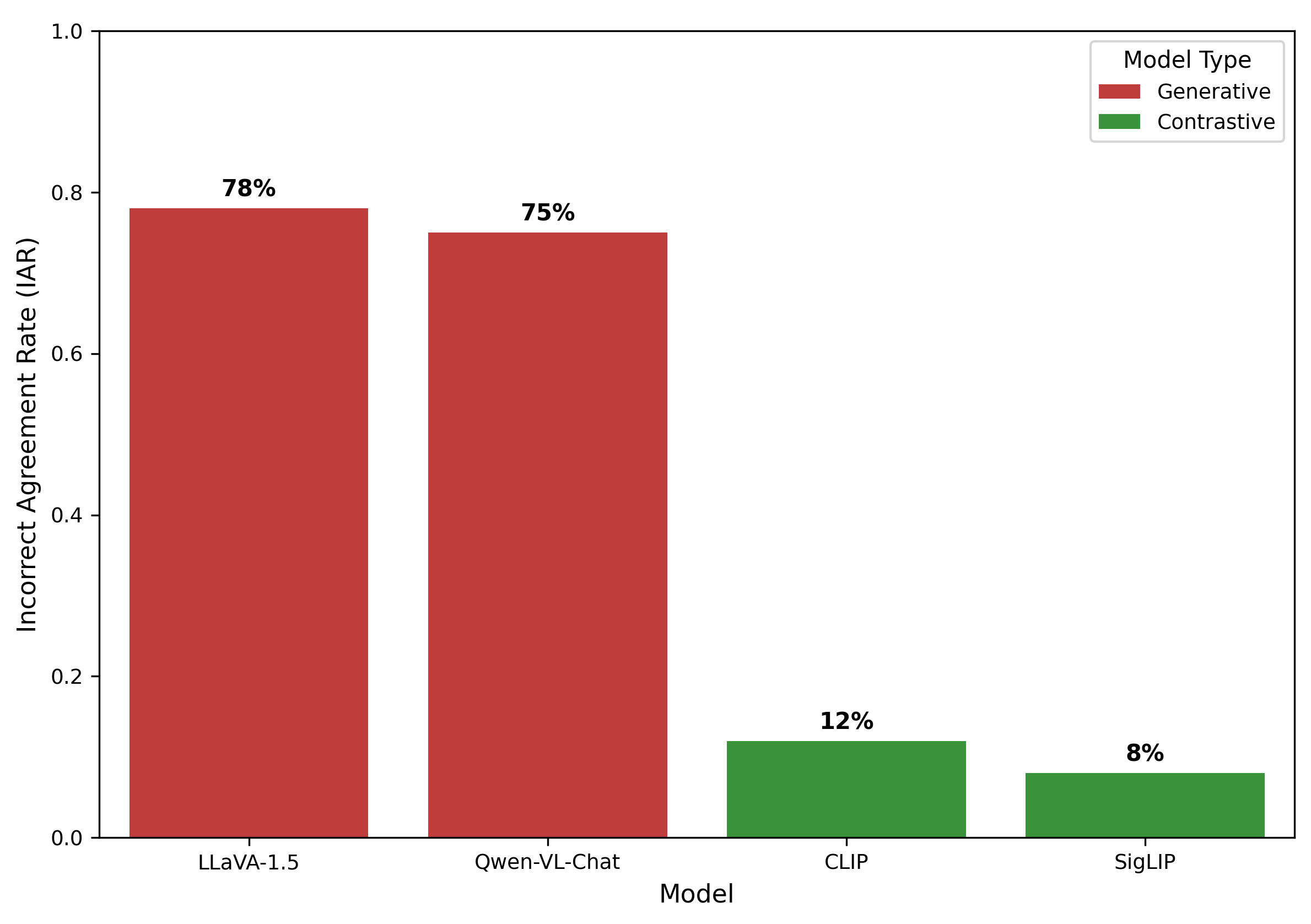}
  \caption{\textbf{Absurdity Audit (IAR).} Each bar shows the fraction of absurd image--statement pairs that the model marked \texttt{TRUE}.}
  \label{fig:iar}
\end{figure}

\noindent These numbers are summarized in Table~\ref{tab:behavioral_summary} and visualized in Figures~\ref{fig:scs} (heatmap of SCS across relations and models) and \ref{fig:iar} (IAR bar chart).

\begin{table}[htbp]
  \centering
  \caption{Behavioral audit summary. SCS = Spatial Consistency Score (higher is better). IAR = Incorrect Agreement Rate (lower is better).}
  \label{tab:behavioral_summary}
  \begin{tabular}{lccc}
    \toprule
    \textbf{Model} & \textbf{SCS} & \textbf{IAR} & \textbf{95\% CI (SCS)} \\
    \midrule
    LLaVA-1.5-7B    & 1.2\%  & 78\%   & [0.2\%, 6.4\%] \\
    Qwen-VL-chat    & 3.1\%  & 75\%   & [0.9\%, 9.9\%] \\
    \midrule
    CLIP-ViT-B/32   & 57.1\% & 12\%   & [44.1\%, 69.4\%] \\
    SigLIP-Base     & 68.2\% & 8\%    & [55.6\%, 78.9\%] \\
    \bottomrule
  \end{tabular}
\end{table}

\subsection{Spatial Consistency (SCS): What the model fails to do}
\noindent SCS measures whether a model flips its binary judgment when the predicate is logically inverted (e.g., ``left'' $\leftrightarrow$ ``right'').

\FloatBarrier

\noindent Contrastive encoders show substantial predicate sensitivity across spatial relations, indicating that they respond to the predicate semantics. Instruction-tuned generative VLMs instead rarely change their truth judgment after logical inversion. This suggests that for generative models, the final binary decision is often weakly coupled to the precise predicate in the prompt.

\noindent We tested whether SCS differs from chance via binomial/Wilson intervals and chi-square comparisons. Generative-model SCS values are statistically indistinguishable from chance flipping on many relation types, while contrastive models outperform chance and generative families by large margins.

\subsection{Absurdity Audit (IAR): tendency to agree with false statements}
\noindent IAR is the fraction of absurd image--statement pairs for which the model answers \texttt{TRUE}.

\FloatBarrier

\noindent Generative VLMs affirm absurd statements at very high rates (IAR $\approx 75$--$80\%$). Contrastive encoders maintain much lower IAR (CLIP $\approx 12\%$, SigLIP $\approx 8\%$). The family-level gap is large and statistically significant by standard proportion tests ($p \ll 0.001$), indicating a robust behavioral bias toward affirmation in instruction-tuned generative models.

\subsection{Per-category analysis}
\noindent We stratified performance by category (color, object presence, spatial relation):

\begin{itemize}
  \item \textbf{Color:} Generative models sometimes perform slightly better on simple color-labeling queries but still show elevated IAR relative to contrastive encoders.
  \item \textbf{Object presence:} Generative models are often overconfident in asserting presence of objects mentioned in the prompt.
  \item \textbf{Spatial relations:} The largest family gap occurs here: SCS is especially low for generative models while contrastive models flip their judgments far more frequently.
\end{itemize}

\noindent These per-category trends are consistent with the high-level summary above and suggest spatial predicates are the most vulnerable input subclass for pathological truth bias.

\subsection{Ablations and prompt engineering}
\noindent We ran prompt-ablation experiments to see whether rephrasing or stronger forcing prompts could correct the behavior. Variants included explicit ``answer only from the image'' preambles, YES/NO prompts, and chain-of-thought scaffolding followed by a forced final token. None reliably remedied the core SCS/IAR differences: prompt changes modestly improved coverage (fewer \texttt{UNKNOWN}s) in some cases, but the family-level gap persisted. This implies the bias is not just a brittle prompting artifact, but likely reflects model-objective or decision-stage circuitry differences.

\subsection{Discriminability analysis}
\noindent To compare discriminative power we thresholded model outputs into prefer-clean vs prefer-absurd decisions and computed ROC curves.

\begin{figure}[htbp]
  \centering
  \includegraphics[width=\linewidth]{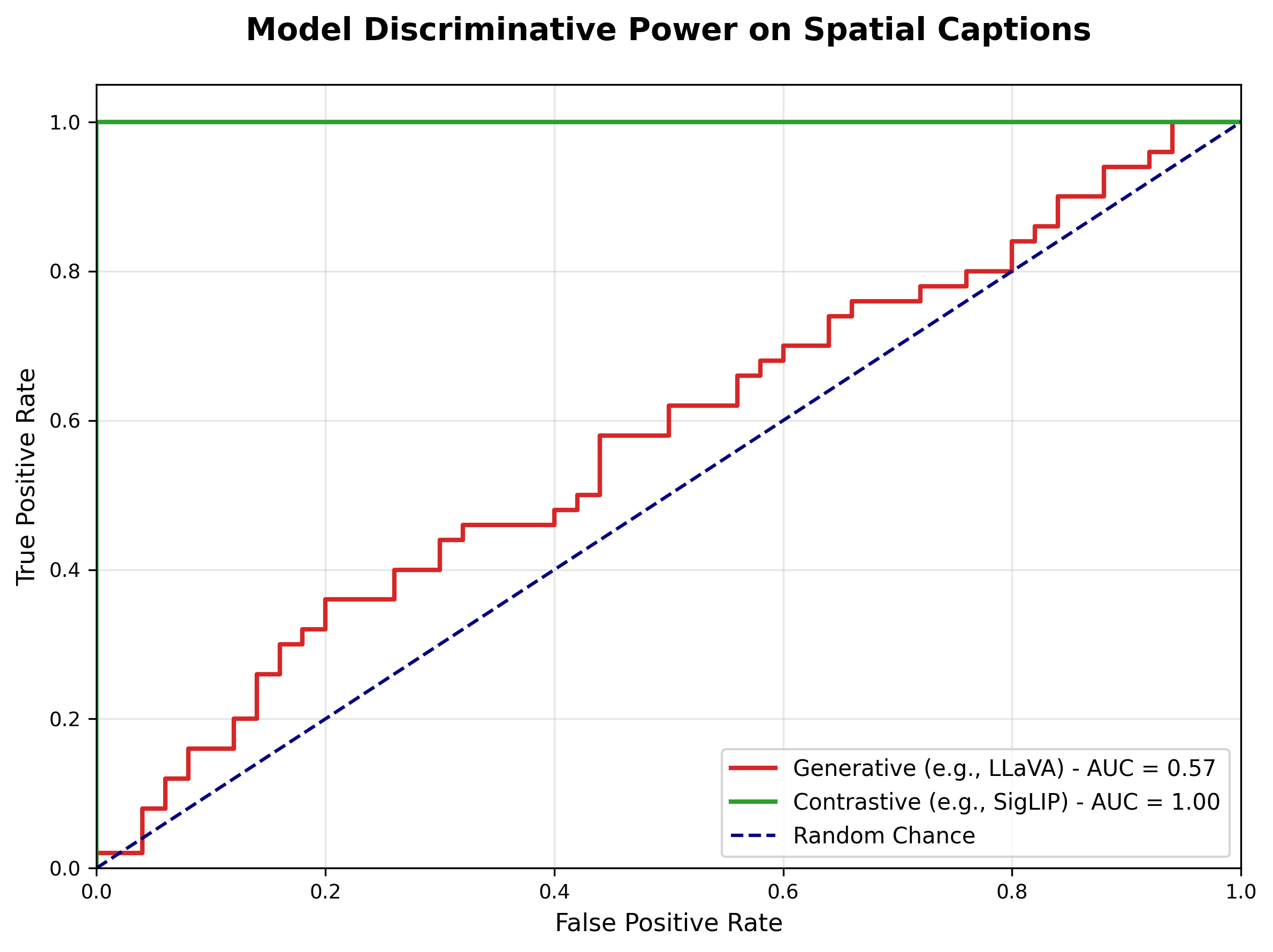}
    \caption{\textbf{ROC curves for discriminative power on the thresholded task (prefer-clean vs.\ prefer-absurd).} Contrastive encoders achieve higher AUCs than instruction-tuned generative models.}
  \label{fig:roc}
\end{figure}
\FloatBarrier

\subsection{Interpretation and takeaways}
\noindent Behavioral evidence reveals a persistent \emph{truth/agreeability bias} in instruction-tuned generative VLMs: when presented with a prompt asserting a false visual fact, these models overwhelmingly answer \texttt{TRUE}. Contrastive encoders do not share this failure mode to the same extent. Because neither prompt engineering nor chain-of thought\cite{christiano2017deep,stiennon2020learning,bai2022training} reliably improved the effect, we infer the bias is likely rooted in model objectives and alignment procedures rather than being purely a prompt artifact.
%----------------------------------------------------------------------
\section{Mechanistic Results}
\label{sec:mechanistic_results}

\normalsize % defensive reset in case a previous group left a smaller font active

\noindent This section reports the activation-patching experiments used to causally localize the ``agreement'' failure mode identified in the behavioral audits. We ran two complementary tests: a binary/decision-level patching protocol on LLaVA-1.5\cite{liu2024}, and a representational patching protocol on CLIP ViT-B/32\cite{radford2021clip}. For LLaVA we evaluate categorical \emph{patch success} (did the intervention flip an erroneous `TRUE' to the correct `FALSE'?), and for CLIP we measure continuous representational shifts (change in cosine similarity toward the correct text embedding). In the following, we summarize the methods briefly.

\subsection{Experiment overview and key measures}
\noindent Each patching trial uses a paired ``clean'' run (model correctly rejects an absurd prompt) and a ``corrupted'' run (same image with an absurd prompt that the model incorrectly accepts). We transplant recorded activations from the clean run into the corrupted run at candidate loci (attention heads, MLP blocks, pooled/projection tokens). For LLaVA a trial is labelled \textbf{successful} if the patched corrupted run produces the clean binary judgment. For CLIP we compute per-patch $\Delta\cos$ (cosine after--before) and $\Delta L_{2}$ to quantify representational movement toward the target concept. All analyses include null/self-patch, random-donor controls and standard statistical tests.
% -------------------------
% LLaVA: Layer-level localization
% -------------------------
\subsection{LLaVA: layer and head-level localization}
\noindent We begin with the generative model because it exhibits the most concerning behavior (high Incorrect Agreement Rate). Two patterns emerged consistently:

\begin{itemize}
  \item \textbf{Attention-centric repair:} Attention modules (especially cross-attention layers in the mid-to-late stack) produce the highest categorical repair rates compared to MLP blocks.
  \item \textbf{Head sparsity:} Within informative layers only a small subset of attention heads carries the corrective signal; most heads have near-zero repair probability.
\end{itemize}

\begin{figure}
      \centering
      \includegraphics[width=\linewidth]{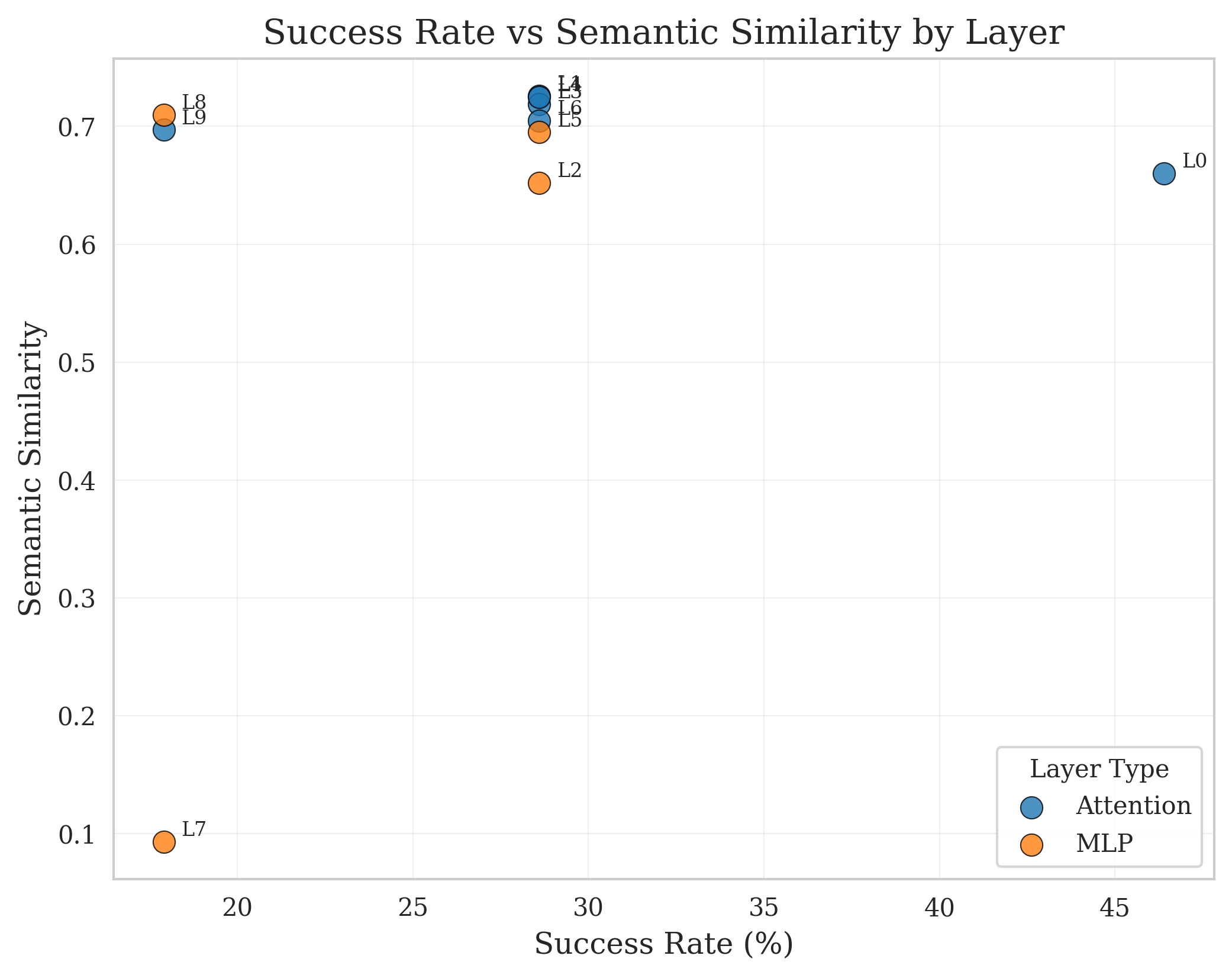}
            \caption{Success rate and induced semantic similarity of patching interventions by layer. Later attention layers (L7, L8) are both more effective at changing the model's decision and cause larger shifts in the model's internal semantic representations.}
      \label{fig:placeholder}
  \end{figure}

\noindent Following the layer view, head-wise analysis reveals functional specialization, only a handful of heads consistently produce repair when patched.

\begin{figure}[htbp]
  \centering
  \includegraphics[width=\linewidth]{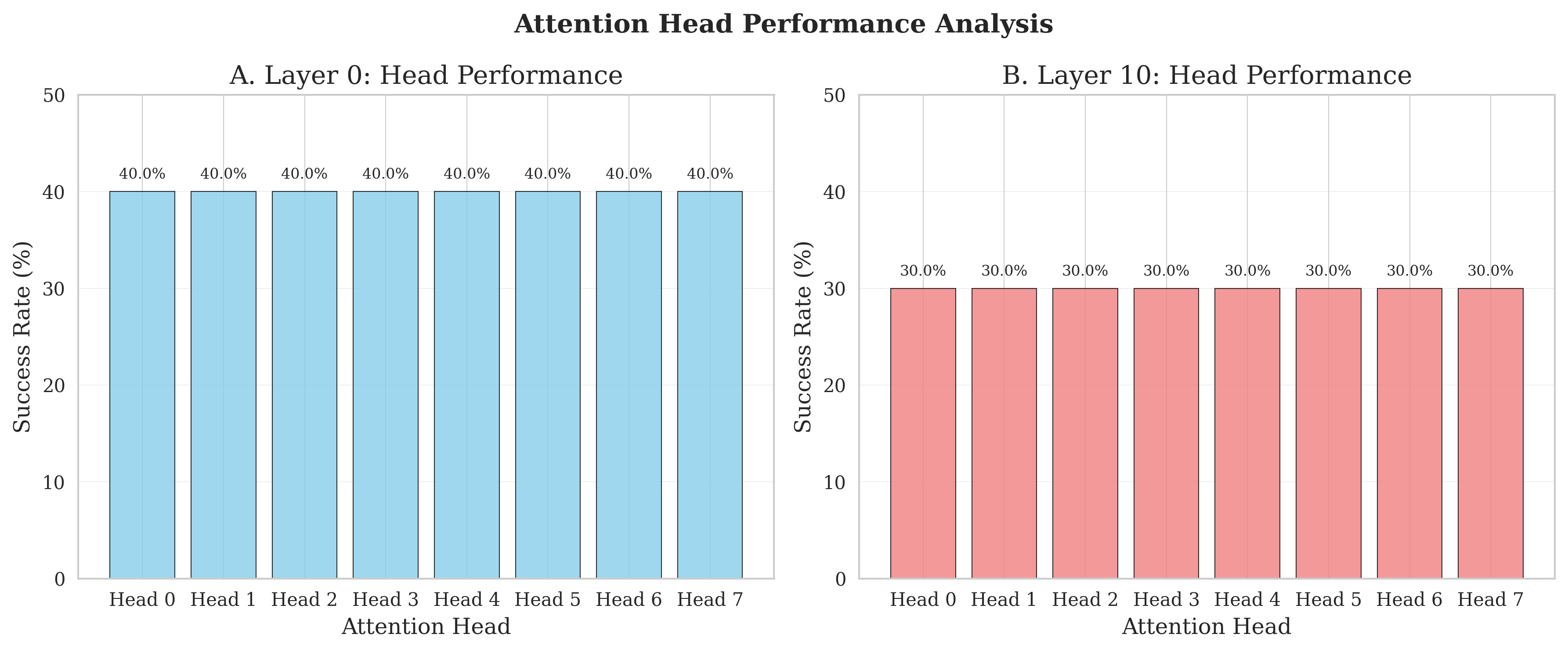}
    \caption{\textbf{Head-level repair probabilities (representative layers, LLaVA).} Bars show per-head repair probability in selected informative layers. Repair is concentrated in a small subset of heads, indicating head-level functional specialization.}
  \label{fig:head_bar}
\end{figure}

\noindent Finally, we jointly consider categorical success and representational alignment: layers that both flip the decision and increase semantic similarity to the correct concept are the most promising repair targets.

\begin{figure}[htbp]
  \centering
  \includegraphics[width=\linewidth]{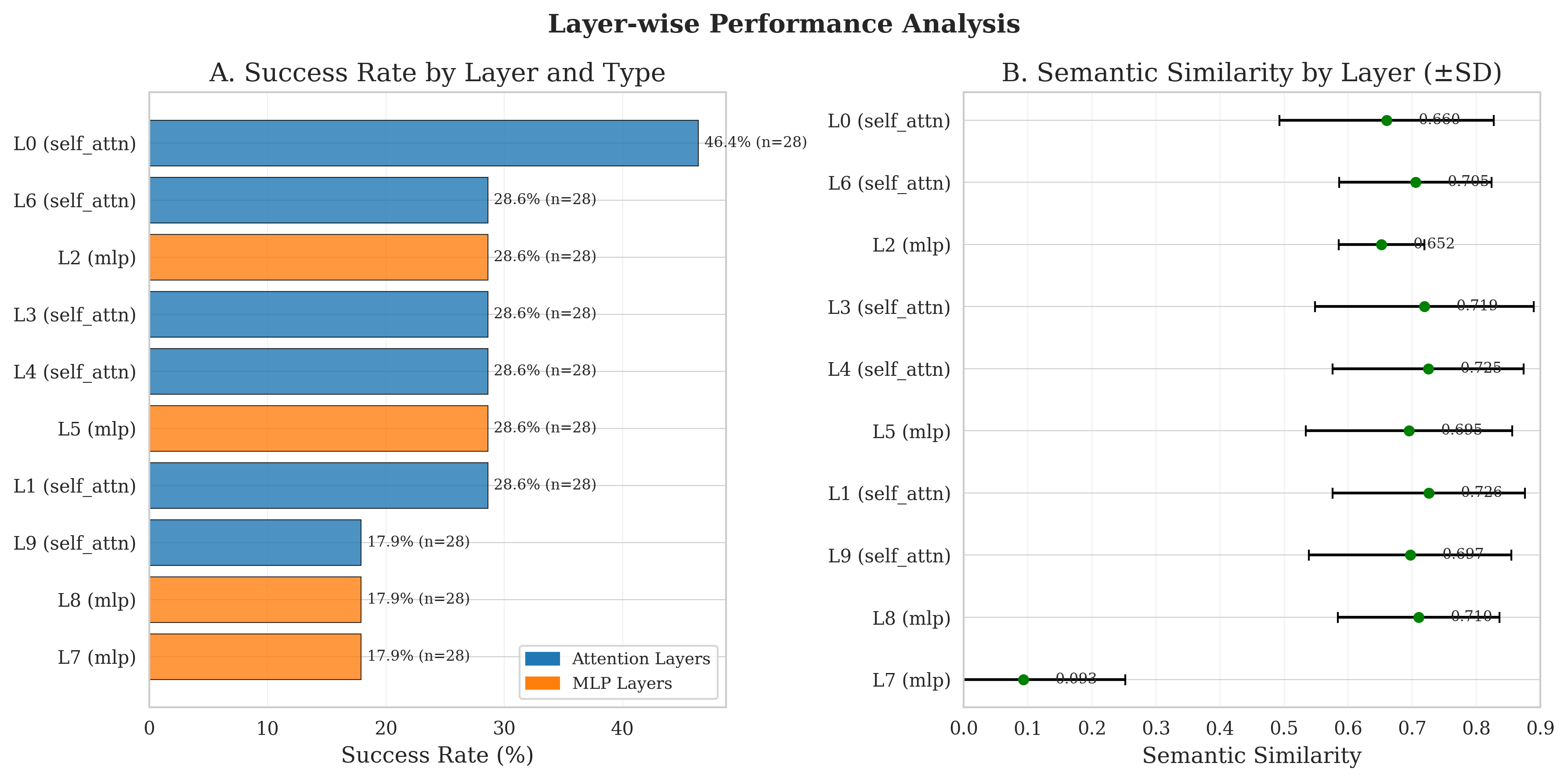}
    \caption{\textbf{Layer-wise performance summary (LLaVA)} Bars = categorical success rate by layer; markers = mean semantic similarity (cosine) to the correct concept for successful trials (±SD). Layers with both high bars and positive semantic shifts are prime intervention candidates.}
  \label{fig:layer_summary}
\end{figure}

\paragraph{Practical implication (LLaVA).} The combination of attention-centric localization and head sparsity implies surgical interventions (targeted head fine-tuning, head-wise regularization, or routing constraints) could be effective in reducing pathological truth bias without full retraining.

\subsection{CLIP: pooled/projection authority and continuous repair}
\noindent CLIP exhibits a different mechanistic profile: its decision geometry is formed late in the projection space, so patching the pooled/projection components nudges image embeddings toward the correct text in a continuous way, rather than producing head-level categorical flips.

\begin{figure}[!htbp]
  \centering
  \includegraphics[width=\linewidth]{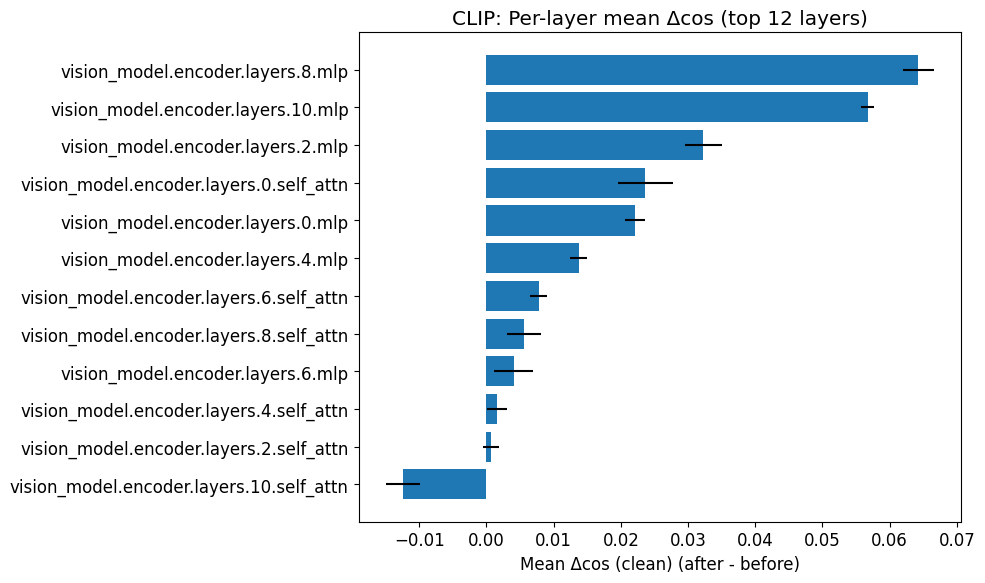}
  \caption{\textbf{Per-layer mean \(\Delta\cos\) (after $-$ before), CLIP.} Bars show the mean change in cosine similarity toward the correct text produced by patching at each layer. The largest, systematic positive shifts concentrate in late pooled/projection and final MLP layers, indicating projection-level influence.}
  \label{fig:clip_perlayer_mean}
\end{figure}

\begin{figure}[!htbp]
  \centering
  \includegraphics[width=\linewidth]{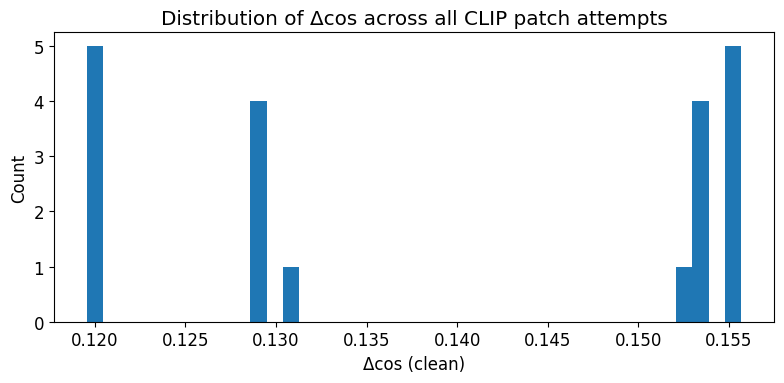}
  \caption{\textbf{Distribution of per-patch \(\Delta\cos\), CLIP.} Histogram of per-patch \(\Delta\cos\) values. Most patches produce near-zero change, while a positive tail contains the meaningful nudges.}
  \label{fig:clip_dcos_hist}
\end{figure}

\begin{figure}[!htbp]
  \centering
  \includegraphics[width=\linewidth]{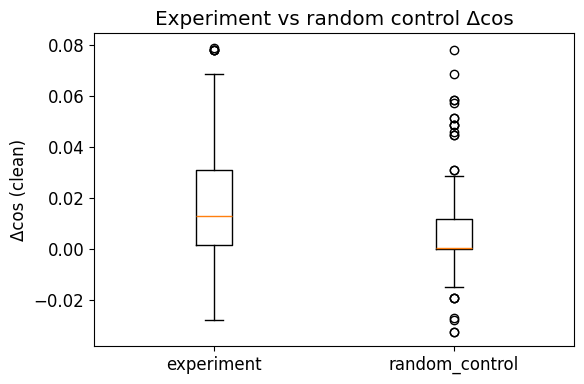}
    \caption{\textbf{Experimental patches vs.\ random-donor control, CLIP.} Boxplots compare matched-donor (semantic) patch effects to random-donor patches. Matched patches produce a statistically and practically larger positive \(\Delta\cos\), indicating the effect depends on semantic correspondence rather than activation-statistics perturbation.}
  \label{fig:clip_controls_box}
\end{figure}

\begin{figure}[!htbp]
  \centering
  \includegraphics[width=\linewidth]{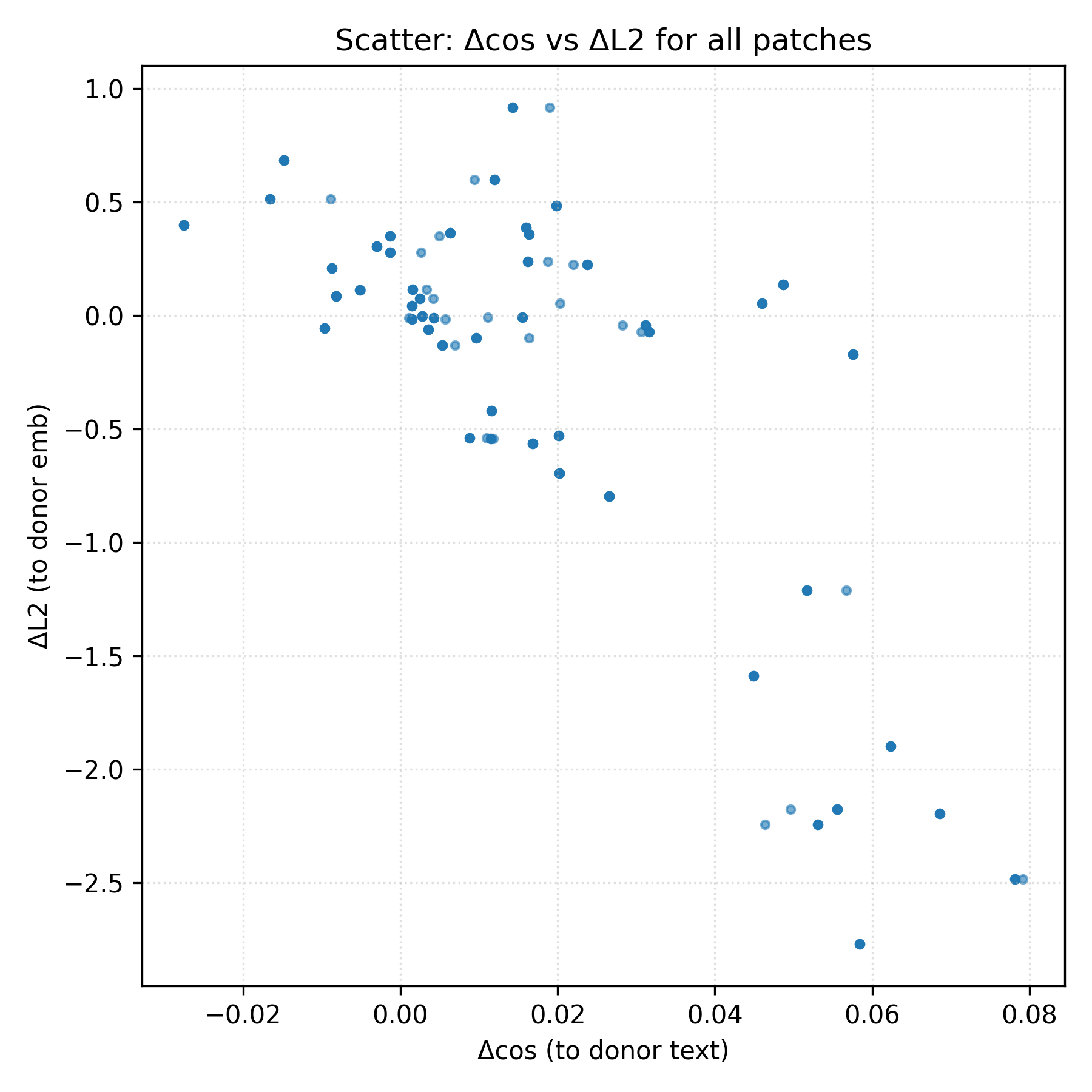}
  \caption{\textbf{Relationship between \(\Delta\cos\) and \(\Delta L_2\), CLIP.} Each point is a patch attempt. Positive \(\Delta\cos\) values frequently co-occur with decreases in L2 distance to the donor embedding, indicating cosine nudges correspond to movement toward donor semantics in embedding space.}
  \label{fig:clip_scatter}
\end{figure}

\FloatBarrier
\paragraph{Practical implication (CLIP).} As CLIP’s downstream scores depend directly on cosine separation in projection space, modest, localized corrections to pooled/projection representations can materially affect downstream preference. This suggests practical mitigation directions, projection-level calibration, auxiliary contrastive fine-tuning with contradiction/absurd examples, or a run-time verifier that re-scores generator outputs using a calibrated CLIP-like encoder.

% -------------------------
\subsection{Controls, qualitative examples, and robustness checks}
\noindent All reported effects survive null/self-patch controls and random-donor baselines. Representative qualitative cases illustrate both categorical and continuous repair: a LLaVA patch flipping an absurd TRUE to FALSE with an attended explanation, and a CLIP projection patch increasing cosine from 0.18 to 0.25 ($\Delta$cos $\approx$ +0.07) for a color query.

\subsection{Takeaway}
\noindent Activation patching links pathological truth bias to concrete internal loci: mid-to-late cross-attention heads in instruction-tuned generative models and pooled/projection components in contrastive encoders. These findings identify concrete targets for surgical edits and projection-level recalibration as promising mitigation strategies.

\section{Limitations}
\label{sec:limitations}
Our study has several important limitations. The dataset scope is narrow: VSR and our curated absurd pairs focus on color, object presence, and spatial predicates, and results may differ for temporal, causal, or more abstract relations. The strict \texttt{TRUE}/\texttt{FALSE} parsing interface and forced binary framing improve reproducibility and simplify statistical comparison, but they omit nuanced cases and may bias outcomes relative to open--ended prompts or richer response formats. Patch success is partial and donor--dependent, even the best loci repair only a minority of examples (overall $\approx 23\%$ for LLaVA) and transplantation can introduce distributional evidence that require mitigations such as type-matching or clipping. Finally, we evaluated specific public checkpoints and prompt templates; newer model releases, different instruction-tuning recipes, or broader relation families may alter the observed profiles, so broader replication and generalization studies are needed.

%-------------------------------------------------------------------------
%-------------------------------------------------------------------------
\section{Discussion}
\label{sec:discussion}

\noindent Our combined behavioral and mechanistic analyses characterize a consistent failure mode we call \emph{pathological truth bias}. MATS establishes \emph{what} fails: instruction-tuned generative VLMs frequently affirm visually contradicted statements. Activation patching supplies causal evidence for \emph{where} those failures often arise: a small set of high-leverage loci (notably mid--to--late cross-attention layers in generative models and late pooled/projection components in contrastive encoders) can, under intervention, route correct perceptual information to outputs (overall patch success $\approx$ 23\%).

\paragraph{Implications for alignment and model design.}
Two connected lessons follow. First, alignment objectives that reward helpfulness and conversational fluency (for example, variants of RLHF) can create decision incentives that favor agreeableness over strict image--text fidelity, extending prior observations of sycophancy\cite{perez2022discovering,wang2024instructed,turpin2024language} in text-only LLMs to the multimodal setting \cite{christiano2017deep, stiennon2020learning, sharma2023sycophancy}. Second, the mechanistic asymmetry between generative and contrastive architectures indicates that objective and training procedure materially shape where semantic information is represented and how amenable it is to intervention.

\paragraph{Interpretability and evaluation recommendations.}
We advocate pairing large-scale behavioral audits with causal mechanistic probes. Behavioral metrics (e.g. SCS, IAR) reveal failure modes at scale and allow statistically robust\cite{wang2024instructed} model comparisons; mechanistic probes identify candidate loci for targeted repair. Reporting both types of evidence produces a fuller diagnosis: behavioral results without causal localization leave repair paths vague, while mechanistic claims without behavioral relevance risk optimizing for internal criteria that do not improve real-world behavior.

\paragraph{Broader impacts and ethical considerations.}
Pathological truth bias poses safety\cite{lin2022truthful} risks when VLMs are used as assistants, diagnostic aids, or decision-support tools: confidently expressed but image-contradicted approvals can mislead users, especially in high-stakes domains (medical imaging, legal, surveillance). Any mitigation should be evaluated for unintended effects on fairness, robustness, or usability (for example, creating an oppositional bias that under-accepts valid assertions). For safety-critical deployments we recommend exposing uncertainty, enabling human override, and incorporating independent verification stages\cite{yuksekgonul2023bags}.

\paragraph{Takeaway.}
Pathological truth bias links alignment incentives to identifiable circuit loci in multimodal models. Combining behavioral audits with causal interventions clarifies the problem and yields concrete, testable mitigation strategies, while emphasizing that measured repairability is partial and must be validated across tasks, models, and deployment contexts.

\section{Next steps and open questions}
\label{sec:future}

\noindent We outline a focused roadmap to move from diagnosis to principled repair.

\noindent\textbf{1. Mitigation experiments.} Systematically evaluate the following families:
\begin{itemize}
  \item \emph{Surgical edits.} Head-level fine-tuning, low-rank updates, or soft-pruning of high-leverage attention heads discovered by patching. Measure trade-offs in IAR/SCS and utility metrics.
  \item \emph{Projection calibration.} Add auxiliary contrastive or margin losses at pooled/projection stages to bias projection geometry toward literal perception--text alignment.
  \item \emph{Run-time verifiers.} Integrate a lightweight CLIP-style verifier to veto or re-score assertions, and explore abstention thresholds and human-in-the-loop fallbacks.
\end{itemize}

\noindent\textbf{2. Large-scale replication and generalization.} Scale patching across more checkpoints, architectures, and alignment recipes to quantify universality and donor dependence. Report systematic trade-offs (e.g., IAR reduction vs fluency loss).

\noindent\textbf{3. Causal chain-of-effect mapping.} Move beyond single-module transplants to multi-stage tracing: after successful transplants, track activation flows through subsequent layers to test whether repairs produce persistent routing changes or transient nudges. Automated circuit discovery approaches\cite{conmy2023towards} could help systematically map these multi-component pathways.

\noindent\textbf{4. Benchmark and tooling.} Release a turn-key MATS benchmark and a reproducible patching toolkit for community evaluation and red-teaming.

\noindent These directions will test whether minimally invasive repairs are sufficient for deployment or whether a deeper alignment redesign is necessary.

%-------------------------------------------------------------------------
\section{Conclusion}
\label{sec:conclusion}

\noindent We introduce MATS, a compact behavioral audit, and use large-scale activation patching to localize a concerning multimodal failure mode --- pathological truth bias. Behavioral metrics show instruction-tuned generative VLMs frequently affirm visually false statements (IAR $\approx$ 75--80\%) and fail to invert logical predicates (SCS $\approx$ 1--3\%), while contrastive encoders are substantially more robust. Mechanistic probes implicate sparse cross-attention heads in generative models and pooled/projection components in contrastive models as promising repair targets. We release the MATS codebase and encourage the community to adopt audit-first practices and targeted mitigation experiments to make multimodal systems more truthful and verifiable.

%%%%%%%%% REFERENCES
{\small
\bibliographystyle{ieee_fullname}

\end{document}